# YOLO-ROC: A High-Precision and Ultra-Lightweight Model for Real-Time Road Damage Detection

Zicheng Lin [1] · Weichao Pan[1]*

**Abstract** Road damage detection is a critical task for ensuring traffic safety and maintaining infrastructure integrity. While deep learning-based detection methods are now widely adopted, they still face two core challenges: first, the inadequate multi-scale feature extraction capabilities of existing networks for diverse targets like cracks and potholes, leading to high miss rates for small-scale damage; and second, the substantial parameter counts and computational demands of mainstream models, which hinder their deployment for efficient, real-time detection in practical applications. To address these issues, this paper proposes a high-precision and lightweight model, YOLO - **R**oad **O**rthogonal **C**ompact (**YOLO-ROC**). We designed a **B**idirectional **M**ulti-scale **S**patial **P**yramid **P**ooling **F**ast (**BMS-SPPF**) module to enhance multi-scale feature extraction and implemented a hierarchical channel compression strategy to reduce computational complexity. The BMS-SPPF module leverages a bidirectional spatial-channel attention mechanism to improve the detection of small targets. Concurrently, the channel compression strategy reduces the parameter count from 3.01M to 0.89M and GFLOPs from 8.1 to 2.6. Experiments on the RDD2022_China_Drone dataset demonstrate that YOLO-ROC achieves a mAP50 of 67.6%, surpassing the baseline YOLOv8n by 2.11%. Notably, the mAP50 for the small-target D40 category improved by 16.8%, and the final model size is only 2.0 MB. Furthermore, the model exhibits excellent generalization performance on the RDD2022_China_Motorbike dataset.

**Keywords** Road Damage Detection · YOLO · Attention Mechanism · Lightweight Model · Multi-scale Feature Extraction · Channel Compression

# 1 Introduction

With the rapid expansion of global economies, road infrastructure has become a cornerstone of societal development and transportation efficiency. The quality and safety of road networks are directly linked to economic vitality and public quality of life. However, road surfaces are perpetually subjected to degradation from traffic loads and environmental erosion, leading to the formation of various types of damage, such as cracks and potholes. These defects pose significant safety risks to both pedestrians and vehicles [1]. Traditional methods for road damage assessment have long relied on manual visual inspection, a process that is not only labor-intensive and time-consuming but also prone to subjective judgment from human inspectors [2]. While automated systems based on specialized sensors have been developed to improve efficiency [3], their widespread adoption is hindered by high equipment costs and a lack of standardized technical protocols. To overcome the efficiency and cost bottlenecks of conventional methods, researchers have increasingly turned to object detection algorithms. Early machine learning approaches, including Support Vector Machines (SVM) and Random Forests, demonstrated limited efficacy due to their heavy reliance on manually engineered features, which failed to generalize across complex and variable road conditions [4]. In recent years, deep learning-based object detection algorithms,

Zicheng Lin
The School of Computer Science and Artificial Intelligence,
Shandong Jianzhu University, Jinan 250101, China
E-mail: 202311102026@stu.sdjzu.edu.cn

Weichao Pan
The School of Computer Science and Artificial Intelligence,
Shandong Jianzhu University, Jinan 250101, China
E-mail: panweichao01@outlook.com
*Corresponding author: Weichao Pan

exemplified by the You Only Look Once (YOLO) series [5], have emerged as the dominant paradigm. Their end-to-end feature learning capabilities have led to significant breakthroughs in both detection accuracy and speed. Nevertheless, two fundamental challenges persist in existing deep learning models for this task. First, they often exhibit insufficient multi-scale feature extraction capabilities, struggling to accurately identify both large-area cracks and minute potholes simultaneously. Second, the substantial number of parameters and high computational complexity of these models make them difficult to deploy on resource-constrained mobile or embedded devices for real-time applications. To address these core issues, current research has primarily explored three optimization directions. (1) **Improving Spatial Pyramid Pooling-Fast (SPPF)**: Models like YOLOv9 [6] have enhanced multi-scale feature fusion without losing spatial information. However, they still suffer from missed or false detections of small targets (e.g., tiny potholes) in complex scenarios with uneven illumination or road debris. (2) **Integrating Attention Mechanisms**: To improve small-object detection accuracy and adapt to diverse road conditions, various attention mechanisms have been introduced. For instance, models like Dal-yolo [7] and YOLO9tr [1] leverage deformable and partial attention blocks, respectively, to dynamically adjust kernel shapes and focus on salient features. Yet, the inclusion of such modules can increase computational load and inference latency. (3) **Model Compression and Lightweight Architectures**: Techniques seen in models like MED-YOLOv8s [8], which employs lightweight backbones inspired by efficient architectures like the MobileNet series [9] and ShuffleNet [10], effectively reduce model parameters and complexity, boosting inference speed. However, this often comes at the cost of reduced accuracy, particularly when targets are occluded, as the simplified models may lack the capacity to handle such complexities. This highlights a persistent trade-off: existing methods struggle to strike an effective balance between precise multi-scale feature extraction and model lightweighting. Therefore, developing an efficient and lightweight road damage detection technology that maintains high accuracy is of critical importance. To resolve the key challenges of inadequate multi-scale feature extraction in complex scenes, excessive model parameters, and the trade-off between lightweight design and accuracy, this paper proposes a high-precision, lightweight, and improved YOLO model, named YOLO-ROC. The objectives of this research are: (1) to design a novel multi-scale feature enhancement module to address the limitations of the traditional SPPF [11] in filtering features under complex interference; (2) to achieve model lightweighting through channel compression and structural optimization, avoiding the accuracy degradation common in existing lightweight methods: and (3) to validate the model's performance on public road damage datasets, demonstrating a superior balance between detection accuracy and computational efficiency. The main contributions of this paper are summarized as follows:

1. **A Novel Bidirectional Multi-scale Module for Feature Enhancement.** We introduce the **B**idirectional **M**ulti-**s**cale **S**patial **P**yramid **P**ooling **F**ast (**BMS-SPPF**) module to address the inadequate multi-scale feature extraction for small and irregular road damage. By integrating a **M**ulti-**S**cale **S**patial **A**ttention (**MSSA**) mechanism and a **M**ulti-**H**ead channel **S**elf-**A**ttention (**MHSA**) unit, BMS-SPPF effectively captures the morphological features of targets like cracks and potholes while suppressing background interference, boosting the mAP50 for the small-target D40 class by 16.8%.

2. **A Hierarchical Compression Strategy for Model Lightweighting.** To tackle the high computational and deployment costs of mainstream models, we propose a hierarchical channel compression strategy. This approach reduces the model's parameter count by 70.4% (from 3.01M to 0.89M) and GFLOPs by 67.9% (from 8.1 to 2.6). Aided by the feature-enhancement capability of BMS-SPPF, the compressed model avoids the typical accuracy trade-off, achieving a 2.11% mAP50 improvement over the baseline YOLOv8n.

3. **A Highly Efficient and Robust Road Damage Detector.** The resulting model, YOLO-ROC, establishes a new state-of-the-art balance between accuracy and efficiency. Validated on the `RDD2022_China_Drone` and RDD2022_China_Motorbike datasets, YOLO-ROC achieves a 67.6% mAP50, outperforming various contemporary YOLO models, while maintaining a minimal footprint of just 0.89M parameters (2.0 MB model size), demonstrating its high practical value for real-time applications.

The remainder of this paper is organized as follows: Section 2 reviews related work on YOLO-based lightweight models, multi-scale feature optimization, and the synergistic design of attention mechanisms and lightweighting. Section 3 details the overall model architecture, the design principles of the BMS-SPPF module, the hierarchical channel compression strategy, and the loss function. Section 4 describes the datasets, experimental setup, and evaluation metrics. Section 5 presents and analyzes the experimental results, including comparative and generalization experiments, and provides visual evidence to validate the effectiveness of our improvements. Finally, Section 6 concludes the paper and discusses future research directions.

## 2 Related Work

In recent years, deep learning-based methods for road damage detection have made significant strides in accuracy and real-time performance. However, a fundamental challenge persists in balancing the trade-off between multi-scale feature extraction in complex scenes and model lightweighting. This section provides a comprehensive review of existing research, organized into three key areas: lightweight detection models, multi-scale feature optimization, and the synergistic design of attention mechanisms with lightweight architectures. We analyze their limitations to contextualize the innovations of our proposed method.

### 2.1 Lightweight Detection Models based on the YOLO Series

The YOLO series has become a dominant framework in road damage detection due to its efficient end-to-end detection capabilities. For instance, YOLOv8n [12] significantly improves inference speed with its decoupled head and anchor-free design, yet its detection accuracy for small targets remains insufficient, particularly in low-resolution imagery. To address information loss in deep networks, YOLOv9t [6] introduced a novel architecture featuring Programmable Gradient Information (PGI) and the Generalized Efficient Layer Aggregation Network (GELAN). While this design reduced the parameter count compared to YOLOv8n, the model still struggles with missed detections of fine cracks under complex lighting conditions. Subsequently, YOLOv10n [13] further explored model efficiency by designing an end-to-end architecture that eliminates the need for Non-Maximum Suppression (NMS). However, its performance on the RDD2022_China_Drone dataset yielded an mAP50 of only 62%, failing to show a significant improvement over YOLOv8n. Although these models have achieved breakthroughs in lightweight design, they generally suffer from a lack of robustness against complex background interference, making it difficult to meet the practical engineering demands for both accuracy and speed.

### 2.2 SPPF Optimization for Multi-Scale Feature Fusion

The SPPF module [11] has become a critical component in object detection models for enhancing multi-scale perception by aggregating contextual features through multi-scale pooling kernels. Researchers have continuously sought to optimize the SPPF structure. For example, YOLOv9 [6] introduced the SPPELAN module, which reconstructs the pooling path with cascaded extended convolutions and residual connections, improving feature fusion efficiency. Similarly, other works have explored dynamic pooling kernel selection mechanisms that adaptively adjust kernel sizes based on the spatial distribution of input feature maps [14]. Similarly, Transformer-based detectors have explored iterative aggregation to efficiently use multi-scale features [15], but these approaches often introduce architectural complexity not suitable for ultra-lightweight models. However, these traditional SPPF optimization methods often lack explicit modeling of semantic correlations between channels. Consequently, the improved SPPF modules still struggle to achieve precise feature filtering and enhancement under complex interference, such as scenarios involving small targets, occlusions, and non-uniform illumination. This limitation is exemplified by the performance of YOLOv9t, which achieved an mAP50 of only 61.4% on the RDD2022_China_Drone dataset.

### 2.3 Synergistic Improvements in Attention and Lightweighting

To enhance model sensitivity to small targets, researchers have attempted to combine attention mechanisms with lightweight designs. For example, Zhao et al. [8] proposed MED-YOLOv8s, which replaces the standard backbone with a lightweight MobileNetv3 and incorporates the ECA attention mechanism to reduce model size while improving accuracy. Similarly, Lan et al. [7] introduced Dal-yolo, which integrates a Deformable Attention mechanism into the backbone to better capture small object details in UAV imagery. These attention strategies, while effective, are part of a broader family of mechanisms, including the foundational Convolutional Block Attention Module (CBAM) [16], which pioneered the combination of spatial and channel attention. Youwai et al. [1] developed YOLO9tr, which adds a partial attention block to the YOLOv9 architecture to enhance feature extraction for pavement damage. In addition to CNN-based approaches, Transformer-based models like EF-RT-DETR [17] have also been proposed for this task, though they present different trade-offs between global feature modeling and computational cost. While these methods show promise, they often face compatibility issues between attention mechanisms and lightweight modules, making it difficult to achieve efficient inference while maintaining high accuracy. For instance, the addition of complex attention modules can increase inference latency, while aggressive model pruning may lead to an increased rate of false positives amidst complex background interference. It is evident that existing methods commonly face a trade-off, making it difficult to achieve efficient inference while maintaining high accuracy. To overcome these limitations, this paper proposes a high-precision and lightweight improved YOLOv8 model, YOLO-ROC, through the synergistic design of a dynamic attention mechanism and structured compression. Specifically, we replace the SPPF module in YOLOv8 with our proposed BMS-SPPF. This module employs the MSSA mechanism to capture the morphological features of cracks and potholes , followed by the CAP module which reduces spatial dimensions to generate a compact feature descriptor. Finally, the MHSA mechanism suppresses background interference to better fuse multi-scale information. Furthermore, we introduce a hierarchical channel compression strategy that halves the maximum number of channels in the backbone network and reduces the repetition count of C2f modules. This compression strategy ensures that the resulting model maintains high detection accuracy while being significantly more lightweight.

## 3 Methodology

This section addresses the challenges inherent in existing YOLOv8 models for road damage detection, namely insufficient multi-scale feature extraction, excessive model parameters, and the imbalance between lightweight design and accuracy. We propose a high-precision, lightweight framework that synergistically integrates dynamic multi-scale feature enhancement with a structured compression strategy to optimize model robustness and deployment efficiency in complex road scenarios. Specifically, we first introduce the BMS-SPPF module, which leverages the MSSA mechanism and the MHSA unit to enhance the morphological feature extraction of small-scale targets, such as cracks and potholes, while suppressing background interference. Second, we propose a hierarchical channel compression strategy that reduces parameter count and computational overhead by decreasing the maximum channel width of the backbone and the repetition frequency of key modules. Finally, the framework is optimized using a dynamic loss function strategy to further improve detection accuracy in challenging conditions like occlusion and non-uniform illumination. The subsequent sections will provide a detailed exposition of the overall model architecture, the design of the BMS-SPPF module, the channel compression strategy, and the loss function.

### 3.1 Model Overview

The YOLOv8 framework, while a potent baseline for general-purpose object detection, exhibits two primary limitations when applied to the specialized task of road damage detection. First, its standard feature extraction pipeline struggles to adequately capture the diverse and often subtle multi-scale features of road damage, leading to missed detections of fine cracks or small potholes. Second, its significant parameter count and computational demands present a substantial barrier to real-time deployment on resource-constrained mobile or edge devices. To surmount these challenges, this paper introduces YOLO-ROC, a high-precision and ultra-lightweight model engineered to strike a synergistic balance between detection accuracy and computational efficiency. As illustrated in Fig 1, the proposed model architecture is founded on the robust YOLOv8 framework but incorporates two principal innovations. First, to address the challenge of multi-scale feature extraction, we replace the standard SPPF module in the backbone with our novel **Bidirectional Multi-scale Spatial Pyramid Pooling Fast (BMS-SPPF)** module. This component is specifically designed to enhance the model's perception of fine-grained spatial details and complex feature interdependencies, which is critical for robustly identifying small or irregularly shaped targets against noisy road backgrounds. Second, to achieve a lightweight design suitable for real-time applications, we implement a **hierarchical channel compression** strategy. This method systematically reduces network width and depth within the backbone and neck, drastically cutting the model's parameter count and computational complexity (GFLOPs). The overall architecture of YOLO-ROC preserves the proven three-part structure of YOLOv8, comprising a Backbone for feature extraction, a Neck for feature fusion, often implemented using structures like Feature Pyramid Networks (FPN) [18], and a Head for detection. Our innovations are strategically concentrated in the backbone to augment its feature representation power while concurrently creating a more compact and efficient model. Crucially, the advanced feature enhancement provided by the BMS-SPPF module compensates for any potential information loss from the compression strategy, enabling our lightweight YOLO-ROC model to maintain and even surpass the detection accuracy of the significantly larger baseline. The specific designs of these core components are detailed in the subsequent sections.

### 3.2 Bidirectional Multi-scale Spatial Pyramid Pooling-Fast (BMS-SPPF)

To overcome the deficiencies of the conventional SPPF module in extracting multi-scale features within complex road environments, we propose the BMS-SPPF.

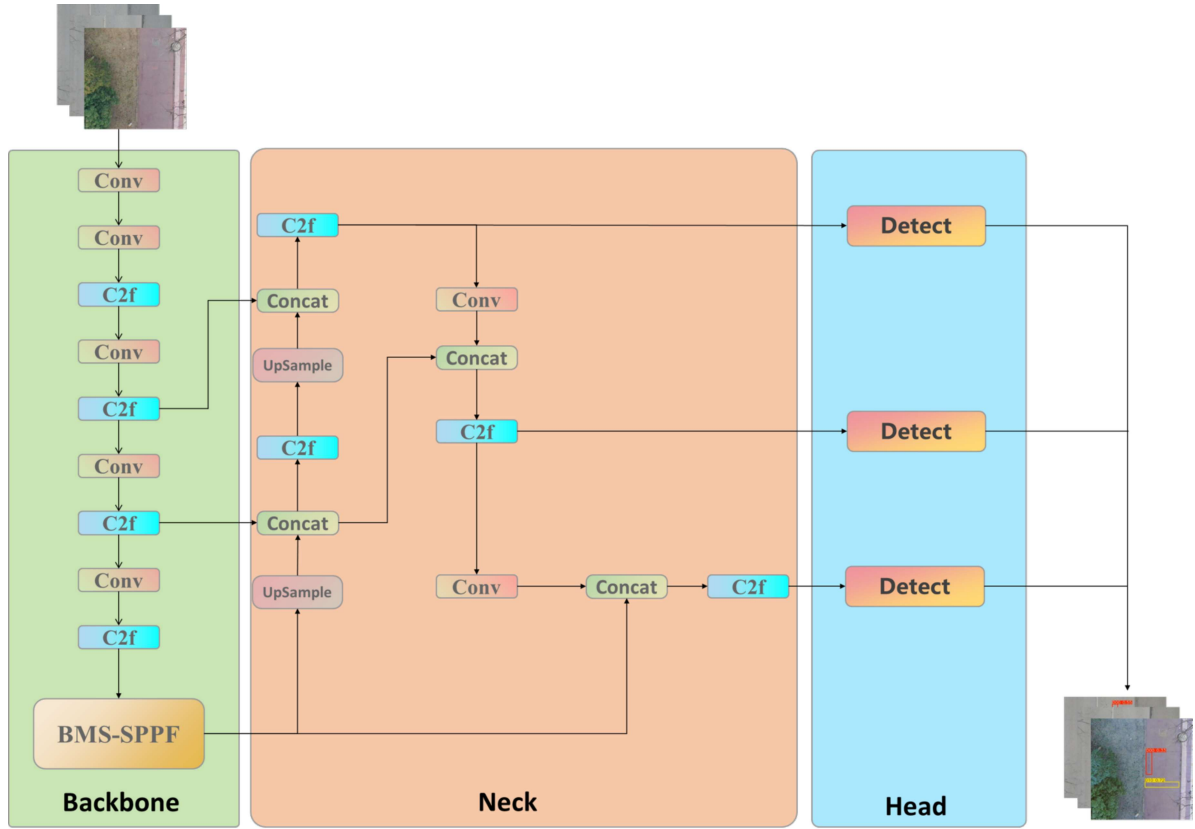

**Fig. 1** Overall architecture of the proposed YOLO-ROC model. The model comprises three main components: Backbone, Neck, and Head. The key innovation, the BMS-SPPF module, is integrated into the Backbone to enhance multi-scale feature extraction.

This module is engineered to dynamically capture diverse morphological features, thereby enhancing the detection accuracy of road damage, particularly fine cracks and small potholes, while maintaining computational efficiency. As depicted in Fig 2, the BMS-SPPF architecture begins with the standard SPPF block, followed by a new attention mechanism composed of three sequential stages: **Multi-Scale Spatial Attention (MSSA)**, a **Channel Attention Preparation (CAP)** module, and **Multi-Head Channel Self-Attention (MHSA)**. Together, these components synergistically enhance multi-scale feature representation and background noise suppression. This section will detail each component, analyzing their individual contributions and collaborative effects.

3.2.1 SPPF Module: Multi-scale Contextual Feature Fusion

To effectively address the significant variations in size and morphology of road damage targets, the model must possess robust multi-scale feature extraction capabilities. To this end, the BMS-SPPF module first retains the efficient SPPF module from YOLOv8 as the initial stage for aggregating multi-scale contextual information. The core advantage of the SPPF module lies in its ability to fuse receptive fields of multiple scales at a low computational cost. Specifically, for an input feature map $\mathbf{X} \in \mathbb{R}^{C \times H \times W}$, the SPPF module concatenates the outputs of several sequential max-pooling layers. The entire process can be formulated as:

$$\mathbf{X}_{\text{sppf}} = \mathcal{F}_{\text{conv}}(\text{Concat}[\mathbf{X}', \mathcal{P}(\mathbf{X}'), \mathcal{P}(\mathcal{P}(\mathbf{X}')), \mathcal{P}(\mathcal{P}(\mathcal{P}(\mathbf{X}')))]) \quad (1)$$

where $\mathbf{X}'$ is the feature map after an initial convolution, $\mathcal{P}(\cdot)$ denotes the max-pooling operation, and $\mathcal{F}_{\text{conv}}(\cdot)$ is the final fusion convolution. This structure aggregates contextual information from different receptive fields, enhancing perception for targets of varying sizes.

3.2.2 Multi-Scale Spatial Attention (MSSA)

To effectively capture the anisotropic features characteristic of road damage, such as the linear structure of cracks, we first introduce a multi-scale spatial attention mechanism. Inspired by seminal works like CBAM [16], which demonstrated the power of decoupling spatial and channel attention, our process begins by selectively enhancing spatially significant features. The process begins by decoupling spatial information along horizontal and vertical axes. For an input feature map $\mathbf{X} \in \mathbb{R}^{C \times H \times W}$, we generate two 1D feature representations by averaging across the orthogonal dimension. This isolates directional context, which can be expressed as:

$$\mathbf{x}_h = \frac{1}{W} \sum_{i=1}^{W} \mathbf{X}_{:,:,i} \in \mathbb{R}^{C \times H} \tag{2}$$

$$\mathbf{x}_w = \frac{1}{H} \sum_{j=1}^{H} \mathbf{X}_{:,j,:} \in \mathbb{R}^{C \times W} \tag{3}$$

where $\mathbf{x}_h$ captures the horizontal context and $\mathbf{x}_w$ captures the vertical context. To endow the model with multi-scale perception, the channels of these 1D feature vectors are divided into four parallel groups. Each group is then processed by a distinct 1D depth-wise separable convolution with a specific kernel size $k \in \{k_1, k_2, k_3, k_4\}$. This strategy allows the model to simultaneously capture both local textural details and long-range spatial dependencies. The outputs are then concatenated, normalized via a group normalization layer, and modulated by a sigmoid gating function to produce the final spatial attention weights, $\mathbf{A}_h \in \mathbb{R}^{C \times H \times 1}$ and $\mathbf{A}_w \in \mathbb{R}^{C \times 1 \times W}$. The original feature map $\mathbf{X}$ is then dynamically reweighted using these attention maps. This enhancement is achieved through element-wise multiplication, as shown in the following equation:

$$\mathbf{X}' = \mathbf{X} \otimes \mathbf{A}_h \otimes \mathbf{A}_w \tag{4}$$

where $\otimes$ denotes element-wise multiplication. This operation enables the model to focus on salient damage regions by explicitly modeling their directional features across multiple scales.

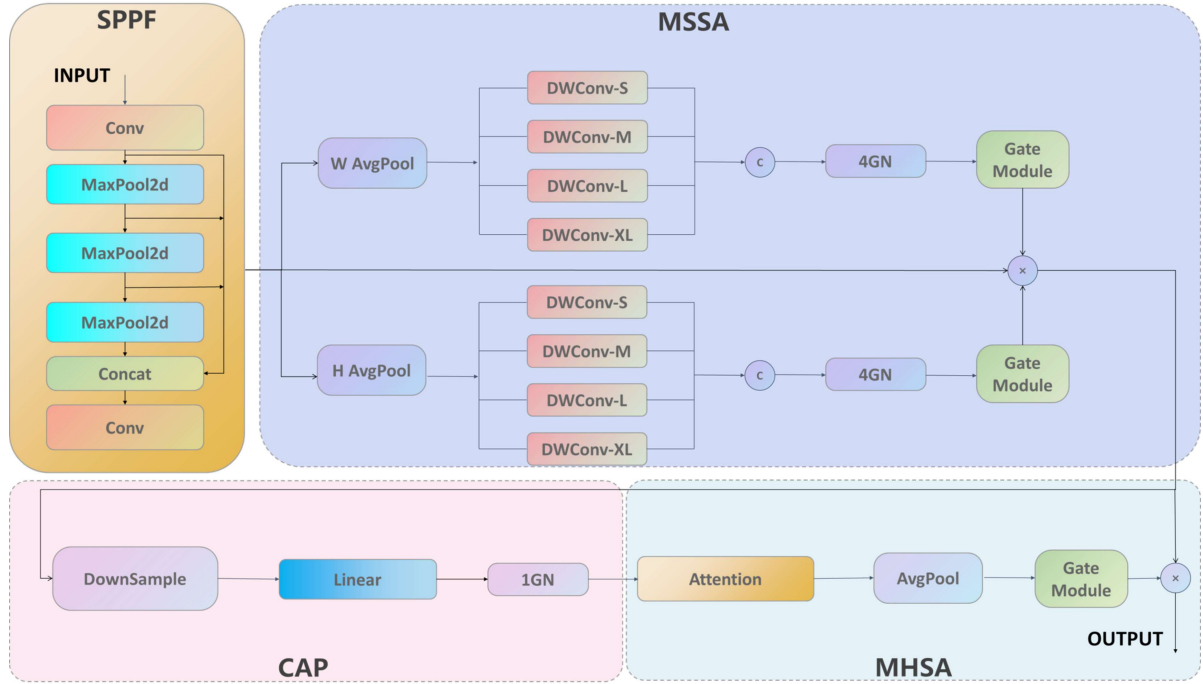

**Fig. 2** Detailed architecture of the proposed Bidirectional Multi-scale Spatial Pyramid Pooling Fast (BMS-SPPF) module. The module's architecture is composed of four sequential stages: SPPF, Multi-Scale Spatial Attention (MSSA), Channel Attention Preparation (CAP), and Multi-Head Self-Attention (MHSA). In the MSSA stage, DWConv-S, -M, -L, and -XL denote 1D depth-wise convolutions with four distinct, progressively increasing kernel sizes (e.g., 3, 5, 7, 9) to capture spatial dependencies at multiple scales. Symbol and block definitions are as follows: "C" denotes the tensor concatenation operation; "X" represents element-wise multiplication; and nGN stands for Group Normalization with n groups.

### 3.2.3 Channel Attention Preparation (CAP)

Following the spatial enhancement from the MSSA module, the feature map $\mathbf{X}'$ undergoes a transformation to produce a compact yet information-rich descriptor for the subsequent channel attention mechanism. This CAP module reduces spatial dimensions through one of two distinct, configurable strategies: lossy pooling or lossless recombination. The first strategy employs a standard spatial pooling function, $\mathcal{F}_{pool}$ (e.g., average pooling), which summarizes local regions and reduces spatial resolution in a lossy manner. The second, more sophisticated strategy is a lossless spatial-to-channel recombination. This operation preserves all feature information by partitioning the feature map $\mathbf{X}' \in \mathbb{R}^{C \times H \times W}$ into non-overlapping spatial blocks of size $s \times s$ and rearranging these blocks into the channel dimension. This transformation, denoted as $\mathcal{F}_{recomb}$, can be formulated as:

$$\mathbf{Y}_t = \mathcal{F}_{recomb}(\mathbf{X}') \in \mathbb{R}^{(C \cdot s^2) \times (H/s) \times (W/s)} \tag{5}$$

This process reduces spatial resolution while encoding the fine-grained spatial details into an expanded channel dimension. A unifying pointwise (1x1) convolution, $\mathcal{F}_{unify}$, is then applied. This step is essential when recombination is used, as it fuses the feature information within the newly expanded channels and restores the channel depth to its original dimension, $C$. If pooling is used, this convolution acts as an identity mapping. This can be expressed as:

$$\mathbf{Y}_p = \mathcal{F}_{unify}(\mathbf{Y}_t) \tag{6}$$

Finally, the resulting feature map is normalized using a group normalization layer, $\mathcal{GN}$, to stabilize its distribution for the attention mechanism:

$$\mathbf{Y}_{norm} = \mathcal{GN}(\mathbf{Y}_p) \tag{7}$$

This completes the feature transformation, yielding a normalized, compact descriptor ready for channel-wise recalibration.

*3.2.4 Multi-Head Channel Self-Attention (MHSA)*

To model the complex interdependencies between channels and dynamically recalibrate feature responses, we employ a multi-head self-attention (MHSA) mechanism. This allows the model to capture more sophisticated, context-dependent channel correlations than methods that rely on simple global pooling. Using the normalized and condensed feature map $\mathbf{Y}_{norm}$ from the previous step, we generate Query ($\mathbf{Q}$), Key ($\mathbf{K}$), and Value ($\mathbf{V}$) projections through parallel, efficient 1x1 depth-wise group convolutions. These projections are then reshaped to separate the attention heads ($h$), allowing the model to jointly attend to information from different representation subspaces. The attention output is calculated using the scaled dot-product attention formula:

$$\text{Attention}(\mathbf{Q}, \mathbf{K}, \mathbf{V}) = \text{softmax}\left(\frac{\mathbf{Q}\mathbf{K}^T}{\sqrt{d_k}}\right)\mathbf{V} \tag{8}$$

where $d_k$ is the dimensionality of each attention head, which serves as a scaling factor to ensure numerical stability. The resulting attention-driven feature map is reshaped, and a final channel attention vector $\mathbf{a}_c$ is generated by computing the mean of this output across its spatial dimensions and applying a Sigmoid gating function. This vector contains weights representing the relative importance of each channel. The final output of the BMS-SPPF module, $\mathbf{X}_{out}$, is produced by applying this channel-wise recalibration to the spatially enhanced feature map $\mathbf{X}'$:

$$\mathbf{X}_{out} = \mathbf{X}' \otimes \mathbf{a}_c \tag{9}$$

This final step ensures that the model amplifies the most informative channel features while attenuating those that are less relevant, leading to a more robust and discriminative final feature representation.

### 3.3 Hierarchical Channel Compression Strategy

To address high parameter counts, we introduce a hierarchical channel compression strategy. This approach follows the principles established by pioneering lightweight networks like MobileNets [9] and ShuffleNet [10], which effectively use channel width reduction to balance efficiency and performance. This involves (1) reducing the maximum channel dimension in the backbone and (2) optimizing module repetition frequency. The original YOLOv8 backbone uses a maximum channel width of 1024. We compress this to 512. We also reduce the repetition count $N^{(l)}$ of the C2f module in shallower layers:

$$N^{(l)} = \begin{cases} 2 & \text{if } l \in \{\text{P3, P4}\} \\ 3 & \text{otherwise} \end{cases} \tag{10}$$

This joint optimization achieves a 67.3% parameter reduction and a 67.9% reduction in GFLOPs. The attention-guided feature screening in BMS-SPPF compensates for information loss from compression. The specific architectural adjustments are detailed in Table 1.

### 3.4 Loss Function

Our model's training is optimized using the established multi-component loss function from the YOLOv8 baseline, ensuring a stable and effective convergence process. This allows the evaluation to focus squarely on the performance gains from our architectural innovations. The total loss, $L_{total}$, is a weighted sum of classification ($L_{class}$), localization ($L_{loc}$), and objectness ($L_{obj}$) losses:

$$L_{total} = \lambda_{class} L_{class} + \lambda_{loc} L_{loc} + \lambda_{obj} L_{obj} \tag{11}$$

where $\lambda$ terms are the corresponding balance coefficients. This composite function effectively handles the dual tasks of classification and localization by integrating a Binary Cross-Entropy (BCE) loss with the Complete-IoU (CIoU) loss for bounding box regression [19], providing a robust foundation for our architectural improvements.

## 4 Experimental Details

### 4.1 Datasets

This study utilizes two subsets from the RDD2022 dataset [20]: RDD2022_China_Drone and RDD2022_China_Motorbike, which contain images of road damage collected in China. The RDD2022_China_Drone subset, used for primary model training and validation, consists of 2,401 training images with 3,068 annotated labels. These images were captured by a hexacopter drone (DJI M600 Pro) along Dongji Avenue in Nanjing, with each image having a resolution of 512×512 pixels. The RDD2022_China_Motorbike subset, used to evaluate the model's generalization capability, contains 2,477 images (1,977 for training, 500 for testing) with 4,650 labels in the training set. These were captured using a smartphone mounted on a motorcycle traveling at approximately 30 km/h around the Jiulonghu Campus of Southeast University, with the same 512×512 pixel resolution. Both datasets are annotated with four primary types of road damage: longitudinal cracks (D00), transverse cracks (D10), alligator cracks (D20), and potholes (D40).

### 4.2 Implementation Details

All experiments were conducted on a workstation running Ubuntu 22.04, equipped with an NVIDIA GeForce RTX 4090D GPU (24GB VRAM). The deep learning framework used was PyTorch 2.1.2 with CUDA 11.8 support, and the programming language was Python 3.10. For training on the RDD2022_China_Drone dataset, we used the Stochastic Gradient Descent (SGD) optimizer with a momentum of 0.937 and a weight decay of 0.0005. The initial learning rate was set to 0.01 and was dynamically adjusted over 300 epochs using a linear learning rate decay scheduler. The batch size was set to 64, and input images were resized to 640×640 pixels. For the RDD2022_China_Motorbike dataset, the Adam optimizer was used, with all other settings remaining consistent. To enhance generalization, mixed-precision training was enabled, and several data augmentation techniques were applied, including Mosaic augmentation, random horizontal flips, HSV color space jitter, random erasing, and RandAugment automatic augmentation.

### 4.3 Evaluation Metrics

To quantitatively assess the performance of the models in the road damage detection task, this study employs a comprehensive set of standard metrics, as established in major survey works [21, 22]. Detection accuracy is measured using Precision (P), Recall (R), and mean Average Precision (mAP). The mAP is reported at two thresholds: mAP50 (IoU threshold of 0.5) and mAP50:95 (averaged over IoU thresholds from 0.5 to 0.95 with a step of 0.05). Model complexity and efficiency are evaluated by the number of Parameters (Params) (in millions, M) and Giga Floating-point Operations per Second (GFLOPs). Deployment cost is measured by Model Size (in megabytes, MB), and inference speed is assessed using Frames Per Second (FPS) to gauge the model's responsiveness for real-time detection tasks.

### 4.4 Comparison Methods

To conduct a comprehensive and objective evaluation of the proposed YOLO-ROC model, we compare its performance against a series of representative baseline and state-of-the-art methods. First, we select mainstream real-time object detection models from the YOLO series, including the YOLOv8n baseline and its subsequent versions YOLOv9t, YOLOv10n. Second, to specifically validate the effectiveness of our core BMS-SPPF module, we introduce advanced SPPF improvement schemes for comparison, such as the SPPELAN module. Furthermore, to evaluate the model's improved capability in detecting small targets, we also include other modified YOLOv8 models that incorporate various attention mechanisms from recent literature [8]. Finally, to verify the competitiveness of YOLO-ROC in its specific application domain, we compare it with other state-of-the-art road damage detection models, such as RT-DSAFDet [23].

## 5 Experiments

To comprehensively evaluate the performance of the proposed YOLO-ROC model, a series of experiments were designed. Section 5.1 validates the model's advantages through comparisons with mainstream models on the RDD2022_China_Drone dataset. Section 5.2 assesses the model's generalization capability on the RDD2022_China_Motorbike dataset. Section 5.3 investigates the impact of the channel compression strategy. Section 5.4 provides a qualitative analysis through visualizations. Finally, Section 5.5 analyzes the sources of error.

Table 1 Simplified architectural comparison highlighting key changes between YOLOv8n and YOLO-ROC.

| Component | Aspect of Change | YOLOv8n (Baseline) | YOLO-ROC (Proposed) |
|---|---|---|---|
| **Backbone** | Width & Depth | Max Channels: 1024; C2f repeats: (3,6,6,3) | **Max Channels: 512;** **C2f repeats: (2,3,3,2)** |
| | Feature Fusion | SPPF Module | **BMS-SPPF Module** (Enhanced) |
| **Head** | Width & Depth | C2f repeats: 3; Channels: 1024/512/256 | **C2f repeats: 2;** **Channels: 512/256/128** |
| **Overall Impact on Efficiency** | | Params: 3.01M; GFLOPs: 8.1 | **Params: 0.89M; GFLOPs: 2.6** |

Table 2 Performance comparison with state-of-the-art models on the RDD2022_China_Drone dataset. Optimizer: SGD.

| Model | P | R | mAP50 | mAP50:95 | Params (M) | GFLOPs | Model Size (MB) |
|---|---|---|---|---|---|---|---|
| *Baseline YOLO Models* | | | | | | | |
| YOLOv8n | 0.749 | 0.593 | 0.662 | 0.397 | 3.01 | 8.1 | 6.3 |
| YOLOv9t | 0.605 | 0.617 | 0.614 | 0.336 | 1.97 | 7.6 | 4.7 |
| YOLOv10n | 0.608 | 0.630 | 0.620 | 0.354 | 2.70 | 8.2 | 5.8 |
| YOLOv11n | 0.635 | 0.632 | 0.664 | 0.382 | 2.58 | 6.3 | 5.5 |
| YOLOv12n | 0.704 | 0.563 | 0.624 | 0.346 | 2.56 | 6.3 | 5.6 |
| *Improved SPPF Variants* | | | | | | | |
| SPPF_improve | 0.720 | 0.665 | 0.711 | 0.409 | 3.07 | 8.1 | 6.4 |
| SPPF_deformab | 0.734 | 0.635 | 0.695 | 0.421 | 6.41 | 10.8 | 13.1 |
| SPPF_uniRepLK | 0.716 | 0.666 | 0.694 | 0.413 | 5.26 | 9.9 | 10.8 |
| SPPELAN | 0.732 | 0.611 | 0.670 | 0.391 | 3.50 | 8.5 | 7.3 |
| *Attention-based YOLOv8 Variants* | | | | | | | |
| yolov8_MSFE | 0.741 | 0.667 | 0.684 | 0.388 | 3.45 | 8.4 | 7.2 |
| yolov8_ASCPA | 0.743 | 0.634 | 0.671 | 0.404 | 3.01 | 8.1 | 6.3 |
| yolov8_C2f_MS | 0.647 | 0.536 | 0.562 | 0.291 | 3.06 | 8.3 | 6.4 |
| *Other SOTA Models* | | | | | | | |
| RT-DSAFDet | 0.747 | 0.577 | 0.664 | 0.396 | 2.74 | 8.0 | 5.9 |
| **YOLO-ROC (Ours)** | **0.738** | **0.608** | **0.676** | **0.397** | **0.89** | **2.6** | **2.0** |

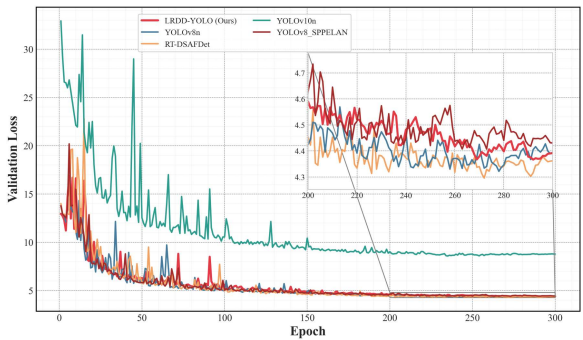

Fig. 3 Comparison of validation loss curves on the RDD2022_China_Drone dataset.

Table 3 Inference speed comparison in Frames Per Second (FPS). 'Total' includes preprocessing, inference, and postprocessing. 'Inference' is the model's forward pass only.

| Model | FPS (Total) | FPS (Inference) |
|---|---|---|
| YOLOv8n | 123.71 | 155.86 |
| YOLOv8_SPPELAN | 132.40 | 163.26 |
| yolov8_MSFE | 102.06 | 118.05 |
| Ours_1024 | 111.04 | 132.57 |
| Ours_256 | 123.28 | 146.42 |
| **YOLO-ROC (Ours)** | **120.19** | **142.74** |

5.1 Comparative Experiments

We conducted comparative experiments against mainstream models on the RDD2022_China_Drone dataset. The results are presented in Table 2 and Table 3. The data indicates that YOLO-ROC achieves an exceptional balance among detection accuracy, model parameters, and computational complexity. Compared to the baseline model YOLOv8n, YOLO-ROC achieves a 2.1% relative improvement in mAP50 (from 66.2% to 67.6%) while significantly reducing the number of parameters from 3.01M to 0.89M (a 70.4% reduction) and GFLOPs from 8.1 to 2.6 (a 67.9% reduction). The model size is compressed from 6.3 MB to just 2.0 MB. Compared to other state-of-the-art road damage detection models like RT-DSAFDet, our model demonstrates superior accuracy with significantly fewer parameters and GFLOPs. This performance enhancement is attributed to the synergistic design of BMS-SPPF, which enhances multi-scale feature extraction, and the hierarchical channel compression strategy, which reduces network redundancy while preserving accuracy. As shown in Fig 3, the validation loss curve of YOLO-ROC exhibits a smoother descent and converges to a lower level compared to other models, indicating optimized convergence characteristics.

**Table 4** Generalization performance comparison on the RDD2022_China_Motorbike dataset. Optimizer: Adam.

| Model | P | R | mAP50 | mAP50:95 | Params (M) | GFLOPs | Model Size (MB) |
|---|---|---|---|---|---|---|---|
| YOLOv8n | 0.843 | 0.812 | 0.869 | 0.538 | 3.01 | 8.1 | 6.3 |
| YOLOv9t | 0.818 | 0.794 | 0.851 | 0.520 | 1.97 | 7.6 | 4.7 |
| YOLOv10n | 0.798 | 0.747 | 0.821 | 0.492 | 2.70 | 8.2 | 5.8 |
| YOLOv11n | 0.819 | 0.782 | 0.854 | 0.514 | 2.58 | 6.3 | 5.5 |
| YOLOv12n | 0.791 | 0.751 | 0.819 | 0.475 | 2.56 | 6.3 | 5.6 |
| SPPF_improve | 0.834 | 0.786 | 0.833 | 0.455 | 3.07 | 8.1 | 6.4 |
| SPPF_deformab | 0.837 | 0.826 | 0.861 | 0.515 | 6.41 | 10.8 | 13.1 |
| SPPF_uniRepLK | 0.850 | 0.843 | 0.886 | 0.557 | 5.26 | 9.9 | 10.8 |
| SPPELAN | 0.854 | 0.823 | 0.863 | 0.498 | 3.50 | 8.5 | 7.3 |
| yolov8_MSFE | 0.855 | 0.816 | 0.876 | 0.540 | 3.45 | 8.4 | 7.2 |
| yolov8_ASCPA | 0.861 | 0.764 | 0.849 | 0.505 | 3.01 | 8.1 | 6.3 |
| yolov8_C2f_MS | 0.852 | 0.802 | 0.851 | 0.514 | 3.06 | 8.3 | 6.4 |
| RT-DSAFDet | 0.881 | 0.806 | 0.890 | 0.562 | 2.74 | 8.0 | 5.9 |
| **YOLO-ROC (Ours)** | **0.839** | **0.810** | **0.866** | **0.518** | **0.89** | **2.6** | **2.0** |

**Table 5** Ablation study on the effect of channel compression. "Ours" denotes the proposed YOLO-ROC model with a maximum channel width of 512. Optimizer: SGD.

| Model | P | R | mAP50 | mAP50:95 | Params (M) | GFLOPs | Model Size (MB) |
|---|---|---|---|---|---|---|---|
| Ours_1024 | 0.678 | 0.668 | 0.691 | 0.407 | 2.91 | 7.6 | 6.1 |
| Ours_256 | 0.713 | 0.498 | 0.578 | 0.307 | 0.35 | 1.3 | 1.0 |
| Ours_128 | 0.529 | 0.405 | 0.436 | 0.221 | 0.20 | 1.0 | 0.6 |
| **YOLO-ROC (Ours)** | **0.738** | **0.608** | **0.676** | **0.397** | **0.89** | **2.6** | **2.0** |

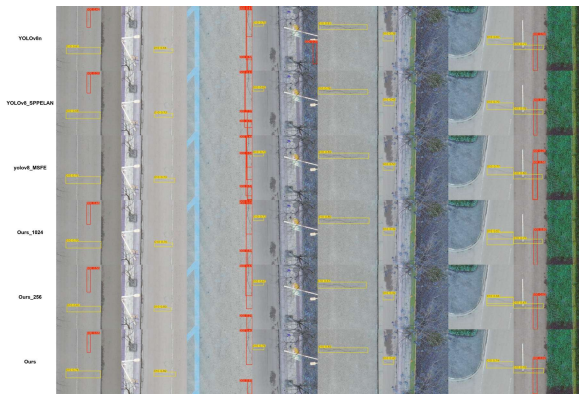

**Fig. 4** Qualitative comparison of detection results on sample images from the RDD2022_China_Drone dataset.

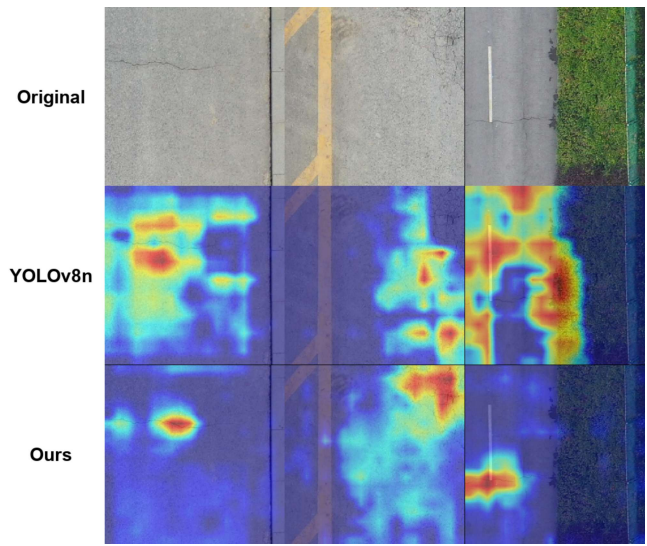

**Fig. 5** Feature heatmap visualization for qualitative analysis.

## 5.2 Generalization Capability Experiments

To validate the generalization capability of YOLO-ROC, we conducted tests on the RDD2022_China_Motorbike dataset, with results in Table 4. Our model achieves an mAP50 of 86.6%, which is competitive with the baseline YOLOv8n (86.9%) but with only 30% of the parameters and GFLOPs. This demonstrates that the model maintains high detection accuracy across different datasets, showcasing its excellent adaptability and efficiency. The model's enhanced generalization is attributed to the BMS-SPPF module and the hierarchical channel compression strategy, which ensure robustness on unseen data while maintaining a lightweight profile.

## 5.3 Ablation Study on Channel Compression

To validate the effectiveness of the hierarchical channel compression strategy, we compared model performance across different maximum channel width settings: 1024, 512 (our proposed model), 256, and 128. The results are detailed in Table 5. The data shows that channel compression significantly reduces computational load. Our proposed model (Ours_512) reduces parameters by 69.4% and GFLOPs by 65.8% compared to the 1024-channel version, with only a minor 1.5% drop in mAP50. This demonstrates high performance retention. In contrast, more aggressive compression (Ours_256 and Ours_128) leads to a sharp drop in accuracy, suggesting that excessive compression impairs the model's expressive power. Therefore, our proposed configuration achieves the optimal balance between detection accuracy and model efficiency.

## 5.4 Visualization Analysis

To intuitively validate performance, we conducted a visualization analysis. A qualitative comparison of detection results is shown in Fig 4, and feature heatmaps are in Fig 5. In the detection results (Fig 4), baseline models like YOLOv8n suffer from false negatives on small cracks. In contrast, our final model (Ours) achieves high-precision localization with confidence scores consistently in the 0.70-0.80 range and more comprehensive coverage of multi-scale targets, thanks to the CAP and MHSA components of the BMS-SPPF module. The heatmap analysis (Fig 5) further corroborates this. The heatmap from YOLOv8n shows dispersed feature responses. In contrast, YOLO-ROC's heatmap displays more focused, high-intensity activation regions precisely at the damage locations. This is attributed to the MSSA mechanism of the BMS-SPPF module, which captures anisotropic features and suppresses background noise.

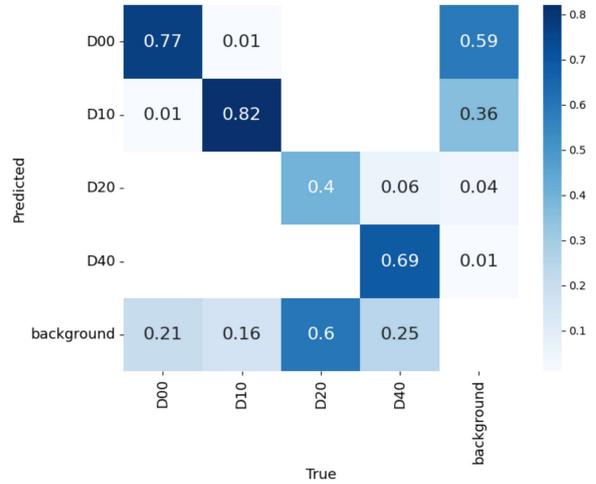

**Fig. 6** Confusion matrix of the proposed YOLO-ROC on the RDD2022_China_Drone dataset.

## 5.5 Error Analysis

Although YOLO-ROC has achieved significant improvements, it still exhibits limitations in certain complex scenarios. The primary source of error stems from deficiencies in detecting minuscule targets, especially under non-uniform illumination or from a high-altitude drone perspective, which can lead to false negatives. The confusion matrix in Fig. 6 reveals the model's classification details. While it performs excellently on D00 (longitudinal cracks) and D10 (transverse cracks), with true positive rates of 77% and 82% respectively, the model still misclassifies a considerable portion of D20 (alligator crack) samples.

## 6 Conclusion

This study proposed YOLO-ROC, a high-precision and lightweight model that optimizes both feature extraction and computational efficiency by integrating a novel BMS-SPPF module and a hierarchical channel compression strategy. On the RDD2022_China_Drone dataset, YOLO-ROC achieves an mAP50 of 67.6%, a 2.11% improvement over the YOLOv8n baseline, while reducing parameters by 70.4% and GFLOPs by 67.9%. The BMS-SPPF module significantly enhanced detection of multi-scale and minuscule targets, while the compression strategy effectively reduced the model's scale. These results validate the superiority of our method in achieving a robust balance between detection accuracy and model efficiency. Although YOLO-ROC has achieved promising results, future work will focus on two aspects: first, introducing an attention guidance mechanism fused with high-level semantic features to address attention dispersion in the BMS-SPPF module under extreme lighting. Second, investigating dynamic differentiable channel pruning to enable the network to adaptively adjust its compression strategy, thereby achieving a better balance between model compression and accuracy preservation.

**Conflict of interest**

The authors declare that they have no conflict of interest.